\documentclass[11pt]{article}

\usepackage[preprint]{acl}

\usepackage{times}
\usepackage{latexsym}
\usepackage{paralist}
\usepackage{amsmath}

\usepackage[T1]{fontenc}

\usepackage[utf8]{inputenc}
\usepackage{subcaption}

\usepackage{microtype}

\usepackage{inconsolata}

\usepackage{graphicx}
\graphicspath{{figs/}{../figs/}}
\usepackage{booktabs}   
\usepackage{amssymb}
\usepackage[table]{xcolor} %
\usepackage{lipsum}
\usepackage{fontawesome5}
\usepackage{array}

\usepackage[toc,page,header]{appendix}
\usepackage{minitoc}
\usepackage{titletoc}

\usepackage{setspace}
\newcommand{\lily}[1]{\textcolor{black}{#1}}

\newcommand{\benchname}{Avalon-ToM-Bench}

\title{Avalon-ToM-Bench: Evaluating Fine-Grained Theory of Mind\\via Asymmetric Game Mechanics}

\author{
  Yen-Shan Chen$^{1,2}$ \quad 
  Yu Chian Duan$^{1,*}$ \quad 
  Chih-En Kuo$^{1,*}$ \quad 
  Jian-Bin Wu$^{1,*}$ \quad 
  Yun-Nung Chen$^1$ \\
  $^1$National Taiwan University \quad
  $^2$CyCraft AI Lab, Taiwan \\
  \texttt{\{r14922018, b12902069\}@csie.ntu.edu.tw}, 
  \texttt{kuochihen@gmail.com}, \\
  \texttt{b12902071@ntu.edu.tw}, 
  \texttt{y.v.chen@ieee.org} \\
  \small{$^*$Equal contribution}
}

\begin{document}
  \maketitle
\begin{abstract}

\lily{Theory of Mind (ToM) is essential for agent interactions, yet existing evaluations either rely on static scenarios that oversimplify mental-state reasoning or interactive settings that provide limited diagnostic insight. We present \textbf{\benchname}, a fine-grained benchmark that operationalizes ToM through the asymmetric-information mechanics of \emph{The Resistance: Avalon}. Rather than evaluating end-to-end gameplay, \benchname{} decomposes ToM into a $2\times2$ taxonomy spanning \emph{epistemic} versus \emph{motivational} reasoning and \emph{inference} versus \emph{action}, using human-crafted, perspective-constrained queries. Benchmarking 28 LLMs reveals three insights: \textbf{First}, models show strong game-rule comprehension but weaker ToM abilities. \textbf{Second}, mechanistic analyses via linear probing and activation steering demonstrate that models frequently represent correct mental-state inferences in their hidden states but fail to express them during generation \lily{(linear probes recover 77--82\% accuracy versus 62--70\% from the models' own chain-of-thought)}, indicating that ToM limitations are primarily a problem of expression rather than representation. \textbf{Third}, dedicated reasoning training yields substantial improvements whereas test-time chain-of-thought provides only marginal gains \lily{($+11.0$ versus $+1.1$ points on average)}, suggesting that robust ToM depends on a learned reasoning policy rather than increased inference-time deliberation.\footnote{Source code available at: \url{https://github.com/yenshan0530/Avalon-ToM-Bench}.}}
\end{abstract}
\section{Introduction}

\begin{figure}[htbp]
    \vspace{-0.5cm}
    \centering \includegraphics[width=0.87\linewidth]{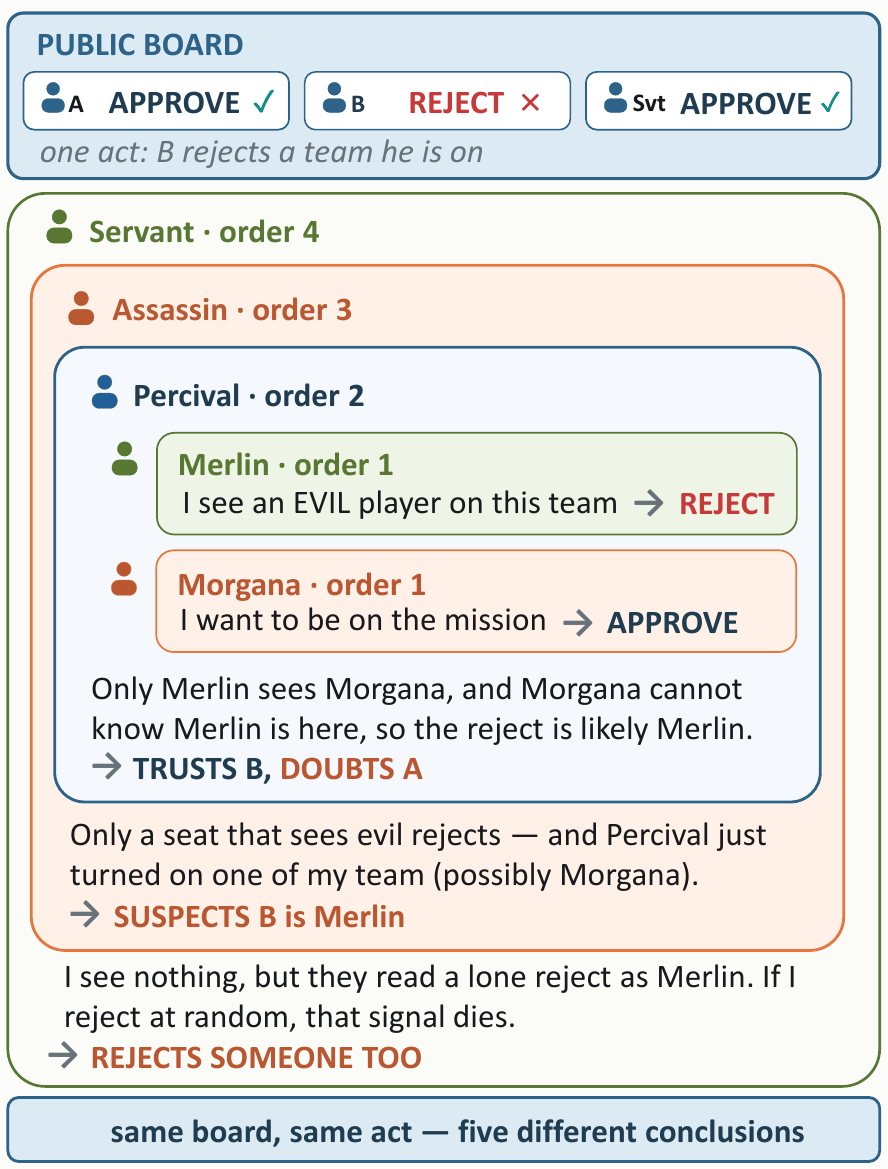}
    \caption{Avalon operationalizes theory of mind. Identical events yield divergent reads, requiring models to deduce hidden perspectives, infer intent from tactics, and shape subsequent actions.}
    \label{fig:teaser}
    \vspace{-1cm}
\end{figure}

\lily{The evaluation of Theory of Mind (ToM)~\cite{premack1978does} in Large Language Models (LLMs) demands a shift from treating the capability as a single, indivisible trait to recognizing it as a spectrum of distinct cognitive abilities. However, prevailing methodologies often miss this nuance.}

\noindent \lily{Traditional benchmarks~\cite{he2023hitom,shinoda2025tomato,gandhi2023understanding} frequently probe superficial patterns through static, overly simplistic scenarios---such as explicitly querying a character's emotional state or visual perspective---stripping away the asymmetric information, deception, and multi-party dynamics inherent to real-world social reasoning. On the other hand, while hidden-role games offer rich, interactive environments~\citep{serrino2019finding,light2023from,light2025strategist,rahimirad2026bayesian}, evaluating models strictly on end-to-end performance (e.g., win rates or action prediction) sacrifices interpretability. Such outcome-driven metrics blur the lines between genuine mind-reading and orthogonal skills like strategic planning, persuasion, or logical memory, making it impossible to isolate true ToM deficits from broader statistical biases.}\\

\lily{To address these limitations, we introduce \textbf{\benchname}, utilizing the mechanics of \emph{The Resistance: Avalon} as a controlled diagnostic environment rather than a gameplay arena. We map ToM across two fundamental axes: the \emph{nature} of the targeted mental state (\textbf{Epistemic} knowledge versus \textbf{Motivational} intent) and its \emph{cognitive execution} (passive \textbf{Inference} versus active \textbf{Action}). This yields four distinct quadrants of asymmetric-information reasoning. To guarantee objective assessment, \benchname{} queries models using highly constrained, human-verified propositions that restrict available evidence exclusively to the designated player's point of view and the public interaction history, penalizing models that rely on omniscient context. \lily{We further verify that these axes are empirically dissociable: under permutation testing, item difficulty loads on the epistemic dimension whereas the benefit of chain-of-thought loads on the orthogonal inference--action dimension, confirming that the taxonomy carves ToM at meaningful joints rather than imposing an arbitrary partition.}}

\lily{By cleanly decoupling ToM from general game-playing heuristics, \textbf{\benchname} is, to the best of our knowledge, the first framework to operationalize mental-state reasoning as a multi-dimensional, testable skill. We benchmark 28 distinct LLMs---encompassing various sizes, reasoning paradigms, and open/closed weights---uncovering three primary insights:}

\begin{compactitem}
    \item \lily{\textbf{Rule Mastery Does Not Equal Social Intelligence:} Models struggle with perspective-constrained ToM (performance $50-60\%$) despite recalling \emph{Avalon's} mechanics near-perfectly (average rule comprehension accuracy $\approx$85\%).}
    \item \lily{\textbf{The Latent Competence Gap:} Linear probing and activation steering reveal that models internally compute the correct deduction but fail to surface it---probes recover 77--82\% accuracy versus 62--70\% from the model's own chain-of-thought, suggesting that ToM failure may be one of expression, not representation.}
    \item \lily{\textbf{Robust ToM Requires Learned Policies, Not Just Deliberation} Test-time chain-of-thought yields volatile, marginal gains ($+1.1$ pts), whereas reasoning-trained models improve consistently ($+11.0$ pts)---robust ToM depends on a learned reasoning policy, not more inference-time tokens.}
\end{compactitem}


\begin{figure*}
    \centering
    \includegraphics[width=\textwidth]{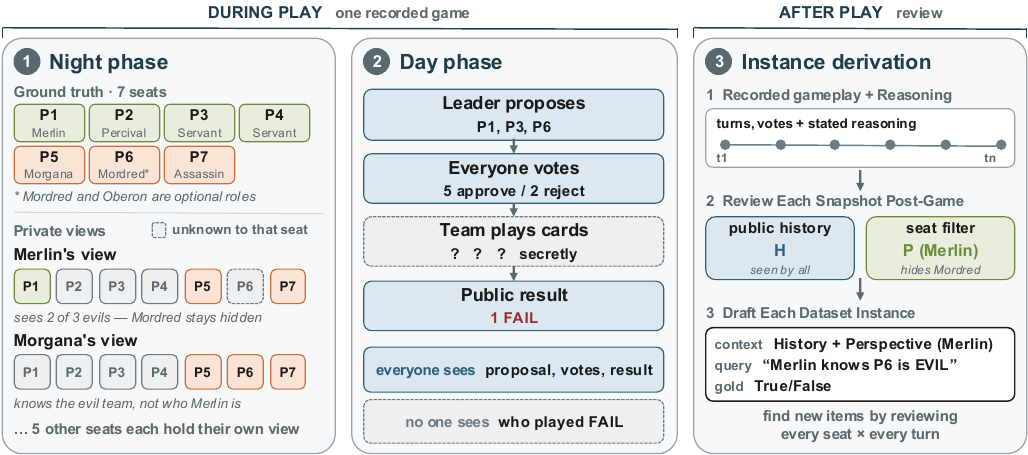}
    \caption{\textbf{Avalon-ToM-Bench overview.} \emph{During play} (1--2), the authors play and
    record a real game: hidden roles give each seat an asymmetric private view of the same public
    events. \emph{After play} (3), each recorded game is reviewed seat-by-seat and turn-by-turn;
    combining the public history $H$ with a seat's role filter $P$ yields one perspective-constrained
    instance (context, query, gold label), forcing strictly perspective-constrained reasoning.}
    \label{fig:avalon-tom-overview}
\end{figure*}

\section{Related Work}
\label{sec:related}



\paragraph{Theory of Mind and its Evaluation in LLMs.}
\lily{Early LLM studies treated ToM~\citep{premack1978does} as a monolithic capability probed by
short, bias-controlled question-answering \citep{le2019revisiting}. Later benchmarks broadened
coverage---conversational asymmetry \citep{kim2023fantom}, long-form narratives
\citep{xu2024opentom}, fine-grained behaviors \citep{chen2024tombench}, and higher-order reasoning
\citep{he2023hitom}---but still frame ToM as passive reading and entangle epistemic knowledge with
motivational intent. Adversarial stress tests expose its brittleness
\citep{shapira2024clever,ullman2023large}, yet \emph{why} models fail, and how to mitigate it,
remains open.}

\paragraph{Social Deduction Games as Reasoning Environments.}
\lily{Hidden-role games are long-standing testbeds for reasoning under information asymmetry, from
pre-LLM game-theoretic agents \citep{serrino2019finding,bakhtin2022human} to a fast-growing line of
LLM \emph{players} \citep{light2023from,wang2024avalon,light2025strategist,xu2024exploring,rahimirad2026bayesian}.
These validate belief inference only through end-to-end game success, conflating it with planning,
persuasion, and memory. Game-based \emph{evaluations} instead target strategic competence
\citep{duan2024gtbench,Lu_2025} or score behavioral alignment to human play
\citep{song2025survivalevaluatingllmssocial,li-etal-2025-inmind}; either way their gold labels are
behavioral rather than the objective correctness of a mental-state deduction.}

\paragraph{Probing and Steering Internal Representations.}
\lily{A complementary interpretability line reads and writes model computation directly: linear
probes recover latent task information \citep{alain2017understanding,belinkov-2022-probing}, models
encode knowledge their generations omit \citep{burns2023discovering,azaria-mitchell-2023-internal},
and activation steering shows such directions are causal \citep{turner2025steering,zou2023transparency}.
We bring these tools to ToM to show that mental-state inferences are internally represented yet
under-expressed---a mechanistic axis unseen in prior ToM benchmarks.} 

\vspace{5pt}
\noindent In Appendix~\ref{app:related}
we discuss related work in detail.

\section{Avalon-ToM-Bench}

\begin{figure*}[t]
    \centering
    \includegraphics[width=\textwidth]{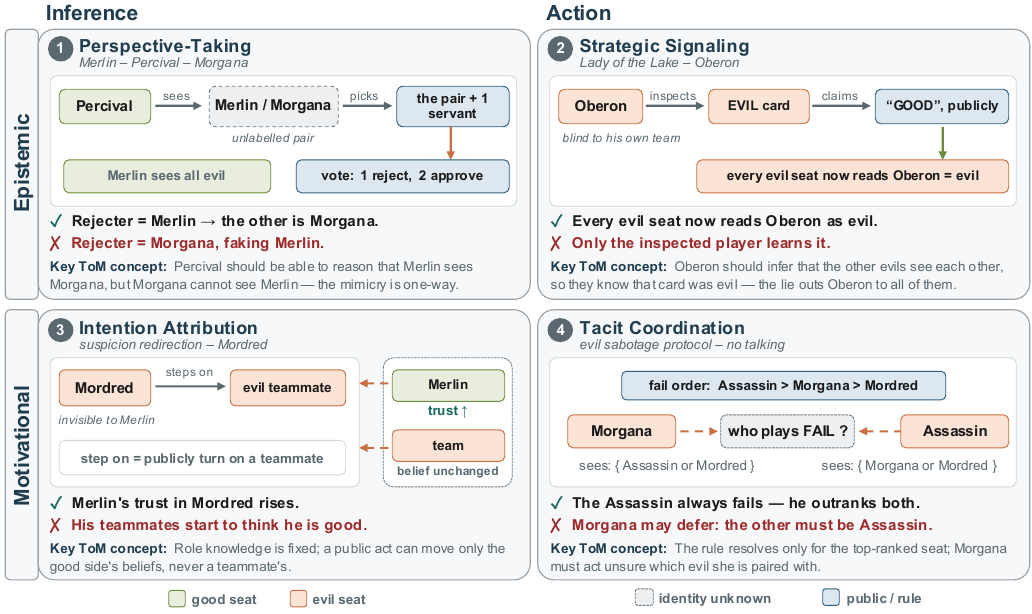}
    \caption{\textbf{The $2\times2$ Theory-of-Mind taxonomy, instantiated in \emph{Avalon}.} The
    \emph{nature of the mental state} (Epistemic vs.\ Motivational) crossed with its \emph{cognitive
    application} (Inference vs.\ Action) yields four dissociable abilities, each realized by a
    naturally occurring Avalon scenario: Perspective-Taking (Merlin--Percival--Morgana), Strategic
    Signaling (Lady of the Lake), Intention Attribution (Suspicion Switching), and Tacit
    Coordination (EVIL Protocol).}
    \label{fig:taxonomy}
\end{figure*}

\lily{Avalon-ToM-Bench treats \emph{The Resistance: Avalon}---a hidden-role game whose secret roles
give each seat asymmetric private information over a shared public history of proposals, votes, and
mission outcomes---not as gameplay to be won but as a measurement instrument. We decompose ToM into
four abilities on a $2\times2$ taxonomy (\S\ref{sec:taxonomy}) and probe each with human-authored,
perspective-constrained binary statements whose labels are objectively verifiable against the rules,
the designated seat's information, and the public history (Figure~\ref{fig:avalon-tom-overview}). The
full ruleset, a self-contained ``Avalon in brief,'' and the role-visibility table are given in
Appendix~\ref{app:benchmark-specification}.}


\subsection{A $2\times2$ Taxonomy of Theory of Mind}
\label{sec:taxonomy}

\lily{We organize ToM along two axes. The first is the \emph{nature of the mental state} being
reasoned about: \textbf{Epistemic}---what a seat (a player's role-defined vantage point) can perceive
or \emph{know} (beliefs, information access)---versus \textbf{Motivational}---what a seat \emph{wants}
or intends (goals, motives). The
second is how that mental state is \emph{engaged}: \textbf{Inference}, where the target mental
state exists from the role and board structure and must be \emph{read off} with no communicative
act involved, versus \textbf{Action}, where a strategic act---a public claim, a sabotage
choice---is itself the object of reasoning and must be produced or interpreted as a deliberate
move. Crossing the two axes yields four dissociable abilities, each of which is naturally
instantiated by a distinct \emph{Avalon} scenario (Figure~\ref{fig:taxonomy}).}

\vspace{5pt}
\noindent \lily{\textbf{Perspective-Taking (Epistemic $\times$ Inference).\ \ } The
\emph{Merlin--Percival--Morgana} structure makes adversarial roles deliberately confusable
(Figure~\ref{fig:taxonomy}); the task is to \emph{read off} what a given seat can legitimately
know under these constraints---a purely epistemic inference with no strategic act.}

\vspace{3pt}
\noindent \lily{\textbf{Strategic Signaling (Epistemic $\times$ Action).\ \ } The \emph{Lady of the
Lake} holder privately learns a player's faction and then makes a public, unverifiable claim; the
evidence is epistemic but the object of reasoning is a communicative \emph{act}---revealing,
concealing, faking, or decoding that claim.}

\vspace{3pt}
\noindent \lily{\textbf{Intention Attribution (Motivational $\times$ Inference).\ \ } \emph{Suspicion
redirection} scenarios reveal no identity directly; the model must abduce a hidden \emph{motive}
from an observed action.}

\vspace{5pt}
\noindent \lily{\textbf{Tacit Coordination (Motivational $\times$ Action).\ \ } The \emph{EVIL sabotage
protocol} (priority Assassin $>$ Morgana $>$ Mordred; Appendix~\ref{app:evil-protocol}) requires EVIL
teammates to converge, \emph{without communication}, on who plays the \textsc{fail} card---a
coordinated \emph{act} that must mesh with teammates' shared intent.}

\vspace{5pt}
\noindent \lily{These placements are design-driven---for each cell we selected the Avalon mechanic that most
naturally instantiates it---and, crucially, not arbitrary: our decomposition analysis
(\S\ref{sec:analysis}) finds that item difficulty loads on the \emph{Epistemic} axis while the
benefit of chain-of-thought loads on the orthogonal \emph{Inference/Action} axis, confirming that
the taxonomy carves ToM at empirically dissociable joints.}

\subsection{Perspective-Constrained Task Formulation}
\label{sec:formulation}

Let $\mathcal{R}$ denote the Avalon ruleset, $G$ the complete game state, $H$ the public
interaction history, and $s$ a candidate inference. Most items designate a player $i$, either
through an explicit perspective instruction $P_i$ or directly in the target query. The
role-visibility rules induce an information projection $I_i=\Pi_i(G)$ containing only what player
$i$ can legitimately know, and the model must evaluate $s$ using evidence restricted to this
projection:
\[
    \hat{y}=M(\mathcal{R},P_i,G,H,s),
    \qquad \hat{y}\in\{\textsc{True},\textsc{False}\}.
\]
Although the model is shown the complete state $G$, valid evidence is confined to $I_i$ and $H$;
it must reconstruct the player's epistemic state rather than treat every stated identity as
accessible knowledge. A role-conditioned statement is labeled \textsc{True} when it is valid under
$\mathcal{R}$, $I_i$, $H$, and any explicit scenario assumptions, and \textsc{False} if it
violates the rules, relies on inaccessible information, or makes an unsupported belief or strategic
inference. The evidence and judgment criteria specific to each dimension are detailed in
Appendix~\ref{app:dimensions}, and one fully worked item per dimension is given in
Appendix~\ref{app:examples}. Statements from the same scenario are evaluated \emph{independently}
rather than as competing multiple-choice options. \lily{A small number of EVIL-Protocol items are
\emph{observer-level}: they test whether a protocol consequence follows from the complete state
without assigning a viewpoint, and are assessed directly against $\mathcal{R}$, $G$, and $H$.}

\vspace{5pt}
\noindent \lily{\textbf{Evaluation.\ \ } Each instance presents the game setup, relevant history, and one
target statement; the gold label and reference rationale are withheld from the model, which
outputs a single \textsc{True}/\textsc{False} judgment. We report accuracy and, because some
diagnostic subsets are class-skewed, macro-averaged $F_1$, which exposes any residual
\textsc{True}/\textsc{False} prior. All items are run zero-shot with a fixed system/user prompt
template (Appendix~\ref{sec:prompts}); decoding regimes and the full model roster are given in
Appendix~\ref{app:model-config}.\lily{
} 
}

\subsection{Data Generation and Validation}
\label{sec:curation}
\definecolor{tomblue}{RGB}{74,144,180}
\definecolor{ptcol}{RGB}{74,144,180}
\definecolor{sscol}{RGB}{74,144,180}
\definecolor{iacol}{RGB}{74,144,180}
\definecolor{tccol}{RGB}{74,144,180}
\definecolor{ntcol}{RGB}{74,144,180}
\newlength{\numw}\settowidth{\numw}{\small\bfseries100.00}\addtolength{\numw}{2pt}
\newcolumntype{R}{>{\raggedleft\arraybackslash}p{\numw}}
\begin{table*}[t]\centering\small
\setlength{\tabcolsep}{3.5pt}\renewcommand{\arraystretch}{1.1}
\resizebox{\textwidth}{!}{%
\begin{tabular}{l *{11}{R}}
\toprule
& & \multicolumn{2}{c}{\parbox[b]{2\numw}{\centering\textbf{Perspective-Taking}}} & \multicolumn{2}{c}{\parbox[b]{2\numw}{\centering\textbf{Strategic Signaling}}} & \multicolumn{2}{c}{\parbox[b]{2\numw}{\centering\textbf{Intention Attribution}}} & \multicolumn{2}{c}{\parbox[b]{2\numw}{\centering\textbf{Tacit Coordination}}} & \multicolumn{2}{c}{\textbf{Average}} \\
& & \multicolumn{2}{c}{\textit{Epistemic}} & \multicolumn{2}{c}{\textit{Epistemic}} & \multicolumn{2}{c}{\textit{Motivational}} & \multicolumn{2}{c}{\textit{Motivational}} & & \\
& & \multicolumn{2}{c}{\textit{Inference}} & \multicolumn{2}{c}{\textit{Action}} & \multicolumn{2}{c}{\textit{Inference}} & \multicolumn{2}{c}{\textit{Action}} & & \\
\cmidrule(lr){3-4}\cmidrule(lr){5-6}\cmidrule(lr){7-8}\cmidrule(lr){9-10}\cmidrule(lr){11-12}
Model & \multicolumn{1}{c}{Rules} & \multicolumn{1}{c}{Acc} & \multicolumn{1}{c}{F1} & \multicolumn{1}{c}{Acc} & \multicolumn{1}{c}{F1} & \multicolumn{1}{c}{Acc} & \multicolumn{1}{c}{F1} & \multicolumn{1}{c}{Acc} & \multicolumn{1}{c}{F1} & \multicolumn{1}{c}{Acc} & \multicolumn{1}{c}{F1} \\
\midrule
\faSquare~Gemini-2.5-Pro & \cellcolor{ntcol!54}\textbf{97.06} & \cellcolor{ptcol!38}79.41 & \cellcolor{ptcol!38}78.50 & \cellcolor{sscol!48}\textbf{87.14} & \cellcolor{sscol!47}\textbf{87.08} & \cellcolor{iacol!55}\textbf{100.00} & \cellcolor{iacol!55}\textbf{100.00} & \cellcolor{tccol!52}\textbf{92.14} & \cellcolor{tccol!51}\textbf{92.13} & \cellcolor{ntcol!50}\textbf{89.67} & \cellcolor{ntcol!50}\textbf{89.43} \\
\faSquare~GPT-5 & \cellcolor{ntcol!54}\textbf{97.06} & \cellcolor{ptcol!42}\textbf{82.35} & \cellcolor{ptcol!42}\textbf{82.29} & \cellcolor{sscol!44}83.57 & \cellcolor{sscol!43}83.12 & \cellcolor{iacol!53}96.67 & \cellcolor{iacol!53}96.67 & \cellcolor{tccol!51}91.43 & \cellcolor{tccol!51}91.38 & \cellcolor{ntcol!49}88.50 & \cellcolor{ntcol!48}88.36 \\
\faSquare~Gemini-2.5-Flash & \cellcolor{ntcol!52}94.12 & \cellcolor{ptcol!35}77.27 & \cellcolor{ptcol!34}77.27 & \cellcolor{sscol!46}86.33 & \cellcolor{sscol!46}86.33 & \cellcolor{iacol!53}95.00 & \cellcolor{iacol!53}94.99 & \cellcolor{tccol!49}88.41 & \cellcolor{tccol!48}88.29 & \cellcolor{ntcol!47}86.75 & \cellcolor{ntcol!47}86.72 \\
\faSquare[regular]~Qwen3.6-35B-A3B & \cellcolor{ntcol!51}91.18 & \cellcolor{ptcol!20}66.18 & \cellcolor{ptcol!19}65.99 & \cellcolor{sscol!39}80.00 & \cellcolor{sscol!38}79.39 & \cellcolor{iacol!53}96.67 & \cellcolor{iacol!53}96.67 & \cellcolor{tccol!49}88.57 & \cellcolor{tccol!49}88.51 & \cellcolor{ntcol!43}82.85 & \cellcolor{ntcol!43}82.64 \\
\faSquare[regular]~Qwen3-4B-Think & \cellcolor{ntcol!51}91.18 & \cellcolor{ptcol!29}72.73 & \cellcolor{ptcol!29}72.50 & \cellcolor{sscol!40}80.58 & \cellcolor{sscol!40}80.38 & \cellcolor{iacol!45}85.00 & \cellcolor{iacol!45}85.00 & \cellcolor{tccol!45}84.89 & \cellcolor{tccol!44}84.84 & \cellcolor{ntcol!41}80.80 & \cellcolor{ntcol!41}80.68 \\
\faSquare~GPT-5-mini & \cellcolor{ntcol!52}94.12 & \cellcolor{ptcol!24}69.12 & \cellcolor{ptcol!24}69.11 & \cellcolor{sscol!33}76.43 & \cellcolor{sscol!32}75.37 & \cellcolor{iacol!47}86.67 & \cellcolor{iacol!47}86.43 & \cellcolor{tccol!49}89.29 & \cellcolor{tccol!49}89.19 & \cellcolor{ntcol!40}80.37 & \cellcolor{ntcol!40}80.03 \\
\faSquare[regular]~Qwen3-8B & \cellcolor{ntcol!54}\textbf{97.06} & \cellcolor{ptcol!23}68.42 & \cellcolor{ptcol!22}67.61 & \cellcolor{sscol!34}77.10 & \cellcolor{sscol!34}77.03 & \cellcolor{iacol!45}85.11 & \cellcolor{iacol!44}84.86 & \cellcolor{tccol!48}87.40 & \cellcolor{tccol!48}87.34 & \cellcolor{ntcol!39}79.51 & \cellcolor{ntcol!38}79.21 \\
\faSquare[regular]~Qwen3-32B & \cellcolor{ntcol!54}\textbf{97.06} & \cellcolor{ptcol!17}64.18 & \cellcolor{ptcol!17}63.98 & \cellcolor{sscol!36}77.94 & \cellcolor{sscol!36}77.90 & \cellcolor{iacol!50}89.83 & \cellcolor{iacol!50}89.76 & \cellcolor{tccol!42}82.01 & \cellcolor{tccol!42}81.98 & \cellcolor{ntcol!38}78.49 & \cellcolor{ntcol!37}78.40 \\
\faSquare[regular]~Qwen3-14B & \cellcolor{ntcol!52}94.12 & \cellcolor{ptcol!14}61.76 & \cellcolor{ptcol!11}60.54 & \cellcolor{sscol!41}80.71 & \cellcolor{sscol!40}80.55 & \cellcolor{iacol!43}83.05 & \cellcolor{iacol!43}83.01 & \cellcolor{tccol!44}83.45 & \cellcolor{tccol!43}83.28 & \cellcolor{ntcol!34}77.25 & \cellcolor{ntcol!33}76.84 \\
\faCircle~GPT-4o-mini & \cellcolor{ntcol!54}\textbf{97.06} & \cellcolor{ptcol!16}63.24 & \cellcolor{ptcol!16}63.16 & \cellcolor{sscol!32}75.71 & \cellcolor{sscol!32}75.63 & \cellcolor{iacol!44}83.33 & \cellcolor{iacol!44}83.31 & \cellcolor{tccol!35}77.86 & \cellcolor{tccol!35}77.80 & \cellcolor{ntcol!32}75.04 & \cellcolor{ntcol!31}74.98 \\
\faSquare[regular]~Qwen3-4B & \cellcolor{ntcol!52}94.12 & \cellcolor{ptcol!7}54.84 & \cellcolor{ptcol!7}54.79 & \cellcolor{sscol!34}76.87 & \cellcolor{sscol!33}76.85 & \cellcolor{iacol!46}86.27 & \cellcolor{iacol!46}86.25 & \cellcolor{tccol!36}78.03 & \cellcolor{tccol!36}78.00 & \cellcolor{ntcol!31}74.00 & \cellcolor{ntcol!31}73.97 \\
\faSquare[regular]~Qwen3.5-9B & \cellcolor{ntcol!48}88.24 & \cellcolor{ptcol!24}68.63 & \cellcolor{ptcol!18}64.71 & \cellcolor{sscol!21}66.94 & \cellcolor{sscol!17}63.70 & \cellcolor{iacol!39}80.00 & \cellcolor{iacol!36}78.02 & \cellcolor{tccol!39}79.70 & \cellcolor{tccol!38}78.61 & \cellcolor{ntcol!31}73.82 & \cellcolor{ntcol!27}71.26 \\
\faCircle[regular]~Phi-4 & \cellcolor{ntcol!51}91.18 & \cellcolor{ptcol!16}63.24 & \cellcolor{ptcol!14}61.84 & \cellcolor{sscol!29}72.86 & \cellcolor{sscol!28}72.17 & \cellcolor{iacol!37}78.33 & \cellcolor{iacol!35}77.58 & \cellcolor{tccol!39}80.00 & \cellcolor{tccol!39}79.85 & \cellcolor{ntcol!30}73.61 & \cellcolor{ntcol!30}72.86 \\
\faCircle[regular]~Gemma-3-27B & \cellcolor{ntcol!52}94.12 & \cellcolor{ptcol!14}61.76 & \cellcolor{ptcol!10}59.52 & \cellcolor{sscol!28}72.14 & \cellcolor{sscol!28}71.73 & \cellcolor{iacol!32}75.00 & \cellcolor{iacol!30}73.77 & \cellcolor{tccol!34}77.14 & \cellcolor{tccol!33}76.84 & \cellcolor{ntcol!27}71.51 & \cellcolor{ntcol!26}70.46 \\
\faSquare[regular]~gpt-oss-120B & \cellcolor{ntcol!54}\textbf{97.06} & \cellcolor{ptcol!16}63.24 & \cellcolor{ptcol!15}62.25 & \cellcolor{sscol!13}61.43 & \cellcolor{sscol!9}58.00 & \cellcolor{iacol!42}81.67 & \cellcolor{iacol!41}81.41 & \cellcolor{tccol!35}77.86 & \cellcolor{tccol!35}77.53 & \cellcolor{ntcol!27}71.05 & \cellcolor{ntcol!25}69.80 \\
\faCircle[regular]~Mistral-7B & \cellcolor{ntcol!46}85.29 & \cellcolor{ptcol!24}69.12 & \cellcolor{ptcol!22}67.54 & \cellcolor{sscol!18}64.29 & \cellcolor{sscol!12}61.11 & \cellcolor{iacol!33}76.67 & \cellcolor{iacol!32}76.00 & \cellcolor{tccol!30}73.57 & \cellcolor{tccol!28}72.21 & \cellcolor{ntcol!27}70.91 & \cellcolor{ntcol!25}69.21 \\
\faCircle[regular]~Llama-3-8B & \cellcolor{ntcol!30}73.53 & \cellcolor{ptcol!16}63.24 & \cellcolor{ptcol!11}60.12 & \cellcolor{sscol!23}67.86 & \cellcolor{sscol!19}65.97 & \cellcolor{iacol!42}81.67 & \cellcolor{iacol!41}81.66 & \cellcolor{tccol!21}67.14 & \cellcolor{tccol!19}65.08 & \cellcolor{ntcol!25}69.98 & \cellcolor{ntcol!23}68.21 \\
\faCircle[regular]~Gemma-3-4B & \cellcolor{ntcol!46}85.29 & \cellcolor{ptcol!22}67.65 & \cellcolor{ptcol!21}67.19 & \cellcolor{sscol!23}67.86 & \cellcolor{sscol!22}67.84 & \cellcolor{iacol!28}71.67 & \cellcolor{iacol!27}71.66 & \cellcolor{tccol!26}70.71 & \cellcolor{tccol!26}70.70 & \cellcolor{ntcol!25}69.47 & \cellcolor{ntcol!25}69.35 \\
\faCircle[regular]~Phi-3.5-mini & \cellcolor{ntcol!51}91.18 & \cellcolor{ptcol!8}55.88 & \cellcolor{ptcol!6}54.46 & \cellcolor{sscol!20}66.43 & \cellcolor{sscol!20}66.04 & \cellcolor{iacol!37}78.33 & \cellcolor{iacol!37}78.33 & \cellcolor{tccol!29}72.86 & \cellcolor{tccol!29}72.40 & \cellcolor{ntcol!23}68.38 & \cellcolor{ntcol!22}67.81 \\
\faCircle[regular]~Llama-3.2-3B & \cellcolor{ntcol!2}26.47 & \cellcolor{ptcol!24}68.89 & \cellcolor{ptcol!22}67.59 & \cellcolor{sscol!10}59.13 & \cellcolor{sscol!10}59.02 & \cellcolor{iacol!37}78.43 & \cellcolor{iacol!37}78.40 & \cellcolor{tccol!9}58.33 & \cellcolor{tccol!7}55.56 & \cellcolor{ntcol!20}66.20 & \cellcolor{ntcol!19}65.14 \\
\faSquare[regular]~Ministral-3-14B & \cellcolor{ntcol!38}79.41 & \cellcolor{ptcol!26}70.59 & \cellcolor{ptcol!25}69.94 & \cellcolor{sscol!18}64.29 & \cellcolor{sscol!12}61.11 & \cellcolor{iacol!7}55.00 & \cellcolor{iacol!4}49.98 & \cellcolor{tccol!26}70.71 & \cellcolor{tccol!24}69.20 & \cellcolor{ntcol!19}65.15 & \cellcolor{ntcol!15}62.56 \\
\faCircle[regular]~Llama-3.1-8B & \cellcolor{ntcol!30}73.53 & \cellcolor{ptcol!5}51.47 & \cellcolor{ptcol!2}40.68 & \cellcolor{sscol!11}60.29 & \cellcolor{sscol!6}54.64 & \cellcolor{iacol!33}76.36 & \cellcolor{iacol!31}74.73 & \cellcolor{tccol!17}64.12 & \cellcolor{tccol!10}59.56 & \cellcolor{ntcol!16}63.06 & \cellcolor{ntcol!8}57.40 \\
\faCircle[regular]~Granite-3.3-8B & \cellcolor{ntcol!51}91.18 & \cellcolor{ptcol!5}51.47 & \cellcolor{ptcol!3}43.89 & \cellcolor{sscol!18}64.29 & \cellcolor{sscol!11}60.37 & \cellcolor{iacol!19}65.00 & \cellcolor{iacol!12}61.10 & \cellcolor{tccol!20}66.43 & \cellcolor{tccol!15}62.56 & \cellcolor{ntcol!14}61.80 & \cellcolor{ntcol!8}56.98 \\
\faSquare[regular]~Qwen3-1.7B & \cellcolor{ntcol!45}85.29 & \cellcolor{ptcol!6}54.10 & \cellcolor{ptcol!5}51.91 & \cellcolor{sscol!21}67.39 & \cellcolor{sscol!21}67.00 & \cellcolor{iacol!5}52.54 & \cellcolor{iacol!5}51.41 & \cellcolor{tccol!29}72.66 & \cellcolor{tccol!28}72.34 & \cellcolor{ntcol!14}61.67 & \cellcolor{ntcol!12}60.67 \\
\faCircle[regular]~Llama-3.2-1B & \cellcolor{ntcol!4}47.06 & \cellcolor{ptcol!6}54.41 & \cellcolor{ptcol!3}42.45 & \cellcolor{sscol!13}61.43 & \cellcolor{sscol!12}61.04 & \cellcolor{iacol!23}68.33 & \cellcolor{iacol!20}66.22 & \cellcolor{tccol!14}62.14 & \cellcolor{tccol!9}57.78 & \cellcolor{ntcol!13}61.58 & \cellcolor{ntcol!8}56.87 \\
\faSquare[regular]~Qwen3.5-2B & \cellcolor{ntcol!46}85.29 & \cellcolor{ptcol!8}55.88 & \cellcolor{ptcol!4}48.33 & \cellcolor{sscol!8}56.43 & \cellcolor{sscol!3}46.22 & \cellcolor{iacol!19}65.00 & \cellcolor{iacol!12}61.10 & \cellcolor{tccol!17}63.50 & \cellcolor{tccol!12}60.63 & \cellcolor{ntcol!11}60.20 & \cellcolor{ntcol!6}54.07 \\
\faSquare[regular]~Qwen3-0.6B & \cellcolor{ntcol!18}64.71 & \cellcolor{ptcol!4}49.25 & \cellcolor{ptcol!4}49.24 & \cellcolor{sscol!15}62.77 & \cellcolor{sscol!15}62.70 & \cellcolor{iacol!13}61.67 & \cellcolor{iacol!13}61.40 & \cellcolor{tccol!11}60.00 & \cellcolor{tccol!10}59.97 & \cellcolor{ntcol!9}58.42 & \cellcolor{ntcol!9}58.33 \\
\faCircle[regular]~OLMo-2-7B & \cellcolor{ntcol!26}70.59 & \cellcolor{ptcol!4}48.53 & \cellcolor{ptcol!2}35.03 & \cellcolor{sscol!6}53.57 & \cellcolor{sscol!3}40.81 & \cellcolor{iacol!8}56.67 & \cellcolor{iacol!3}46.65 & \cellcolor{tccol!16}62.86 & \cellcolor{tccol!9}57.91 & \cellcolor{ntcol!7}55.41 & \cellcolor{ntcol!3}45.10 \\
\midrule
\textbf{Average} & \cellcolor{ntcol!44}84.77 & \cellcolor{ptcol!17}63.46 & \cellcolor{ptcol!12}60.80 & \cellcolor{sscol!26}70.42 & \cellcolor{sscol!24}68.54 & \cellcolor{iacol!36}78.21 & \cellcolor{iacol!34}77.01 & \cellcolor{tccol!32}75.83 & \cellcolor{tccol!31}74.70 & \cellcolor{ntcol!28}71.98 & \cellcolor{ntcol!26}70.26 \\
\bottomrule
\end{tabular}}
\caption{Model performance on the \textbf{full balanced set} (408 items, 50\% True per ability). For each ToM ability we report accuracy (\emph{Acc}) and \textbf{macro-F1} (\emph{F1}). Marker: \faSquare\,/\,\faCircle\ $=$ reasoning\,/\,non-reasoning; filled\,/\,outline $=$ proprietary\,/\,open-weight. \emph{Rules}: full game-rules exam. \emph{Average} is the macro-average over the four abilities. Darker shading $=$ higher; best per column in \textbf{bold}; rows sorted by average accuracy.}
\label{tab:main_results}
\end{table*}

\lily{The benchmark is built from \emph{real games} rather than synthetic vignettes: the authors play
and record full games and then, reviewing each recorded snapshot seat-by-seat and turn-by-turn,
draft and independently cross-validate perspective-constrained statements against the rules, the
seat's role-filtered information, and the public history (Figure~\ref{fig:avalon-tom-overview}; full
protocol in Appendix~\ref{app:curation}). In total it comprises 408 independently evaluated binary
statements across 97 scenarios and the four dimensions, near-balanced at roughly 50\% \textsc{True},
plus a 34-item rule-recall control set---not counted among the 408---that isolates basic rule recall
from perspective-sensitive reasoning (per-dimension statistics in Appendix~\ref{app:stats}).}

\section{Experiments}
\label{sec:experiments}

We evaluate 28 models spanning proprietary and open-weight families, reasoning-trained and
standard instruction-tuned variants, and sizes from 0.6B up. Each is prompted zero-shot with a
fixed template---game setup, public history, and one perspective-constrained statement---and
returns a \textsc{True}/\textsc{False} judgment (\S\ref{sec:formulation}). We report accuracy and
macro-$F_1$ per ability on the full balanced set (408 items, 50\% \textsc{True} per ability), so a
residual \textsc{True}/\textsc{False} prior surfaces as an Acc--$F_1$ gap. A companion \emph{rules
exam} tests recall of Avalon's mechanics, separating game knowledge from reasoning about minds.

\subsection{Models Read the Rules but Not the Mind}
\label{sec:main-results}

The leaderboard in Table~\ref{tab:main_results} reveals three findings:

\paragraph{A wide, structured spread.} Proprietary reasoning models lead (Gemini-2.5-Pro 89.7,
GPT-5 88.5, Gemini-2.5-Flash 86.8), the weakest open models fall to the mid-50s, and class
means track capability: 84.1 (proprietary), 71.9 (open reasoning), 66.5 (open
non-reasoning). Reasoning training and scale both help, yet no open model reaches the
proprietary frontier and even the best leave headroom on the harder abilities.

\paragraph{Rule mastery is not social intelligence.} Almost every model answers the rules exam
well (mean $\approx$83.3\%; most $\geq$88\%) yet scores far lower on ToM items that apply those
same rules from a constrained point of view. The failures are thus deficits of mental-state
reasoning, not missing domain knowledge---and a benchmark that did not separate the two would
conflate them.

\paragraph{The abilities are not one skill.} Averaged over models, accuracy orders as Intention
Attribution (78.2) $>$ Tacit Coordination (75.8) $>$ Strategic Signaling (70.4) $>$
Perspective-Taking (63.5), with Perspective-Taking consistently hardest and Intention
Attribution easiest across the roster. ToM here behaves as dissociable abilities rather than a
single trait---which we make precise next.

\subsection{The Taxonomy Carves ToM at Dissociable Joints}
\label{sec:analysis}

The per-ability ordering could reflect an arbitrary difficulty ranking. To test whether
the two axes of our taxonomy (\S\ref{sec:taxonomy}) truly capture distinct structure, we
decompose two per-cell quantities over the $2\times2$ grid---item \emph{difficulty} (mean
accuracy) and the \emph{benefit of chain-of-thought} (\S\ref{sec:cot})---into main effects for
the Epistemic/Motivational axis and the Inference/Action axis, and assess each effect with a
permutation test that shuffles cell labels.

\begin{table}[t]
\centering
\small
\setlength{\tabcolsep}{6pt}
\renewcommand{\arraystretch}{1.2}
\resizebox{\columnwidth}{!}{%
\begin{tabular}{l cc cc}
\toprule
 & \multicolumn{2}{c}{\textbf{Difficulty}} & \multicolumn{2}{c}{\textbf{CoT benefit}} \\
\cmidrule(lr){2-3}\cmidrule(lr){4-5}
Effect & pp & $p$ & pp & $p$ \\
\midrule
Epistemic vs.\ Motivational & $-8.5$ & \textbf{$<$0.001} & $-1.2$ & 0.71 \\
Inference vs.\ Action       & $-3.2$ & 0.14          & $-7.9$ & \textbf{0.008} \\
Interaction                 & $-9.3$ & \textbf{0.027} & $-4.8$ & 0.45 \\
\bottomrule
\end{tabular}%
}
\caption{\textbf{The two taxonomy axes control different things.} Per-axis main effects on
difficulty and CoT benefit (full set; effect in pp, permutation-test $p$).}
\label{tab:dissociation}
\end{table}

The two quantities load on \emph{orthogonal} axes (Table~\ref{tab:dissociation}). Difficulty
is driven by the Epistemic/Motivational axis ($-8.5$ pp, $p<0.001$): epistemic abilities
(Perspective-Taking, Strategic Signaling), which demand reconstructing what another seat can
perceive, are reliably harder, while the Inference/Action axis has no significant effect on
difficulty ($-3.2$ pp, $p=0.14$). The benefit of chain-of-thought shows the mirror-image
pattern: it is governed by the Inference/Action axis ($-7.9$ pp, $p=0.008$)---deliberation
helps the \emph{Action} abilities, where a strategic move must be produced or decoded, more
than the \emph{Inference} abilities---while the Epistemic axis has no significant effect
($-1.2$ pp, $p=0.71$). This double dissociation confirms that the taxonomy partitions ToM
along meaningful, empirically separable dimensions rather than imposing an arbitrary grouping:
\emph{what} makes an item hard and \emph{what} deliberation can fix are controlled by
different axes of the design. A significant difficulty interaction ($-9.3$ pp, $p=0.027$)
further indicates that Perspective-Taking---epistemic \emph{and} inference---is harder than the
sum of the two main effects would predict, consistent with it being the single hardest cell.

\section{Ablation Studies}

Beyond ranking models, \benchname{} is designed to localize \emph{where} and \emph{why} Theory
of Mind breaks. We report two analyses, conducted on the open-weight models whose internal
states we can access. The first (\S\ref{sec:repr}) shows that the correct answer is present in a
model's internal representation yet under-expressed at its output: ToM failure is largely a
problem of \emph{expression, not representation}. The second (\S\ref{sec:cot}) shows that
test-time chain-of-thought does little to close this gap, whereas dedicated reasoning
\emph{training} does---implicating the learned reasoning policy rather than the amount of
inference-time deliberation.

\subsection{Internal Representations as Stronger Predictors of ToM}
\label{sec:repr}
If low accuracy reflected an inability to \emph{represent} the answer, the answer should be as
absent from a model's internal states as from its output. We find the opposite through three distinct perspectives: 
\begin{figure*}[!ht]
  \centering
  \begin{subfigure}[b]{0.63\textwidth}
    \centering
    \includegraphics[width=\linewidth]{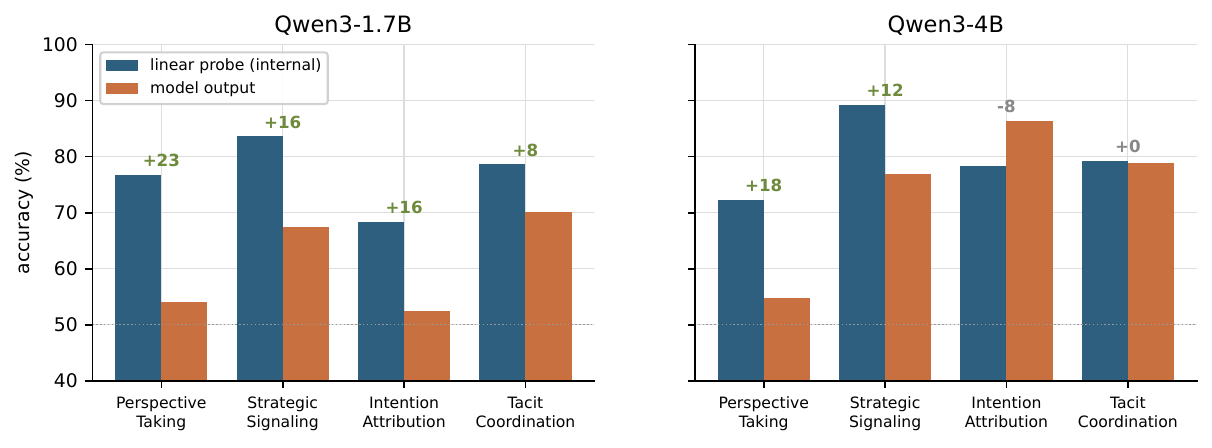}
    \caption{Probe\,$-$\,output gap, per ability.}
    \label{fig:probe_gap}
  \end{subfigure}\hfill
  \begin{subfigure}[b]{0.34\textwidth}
    \centering
    \includegraphics[width=0.95\linewidth]{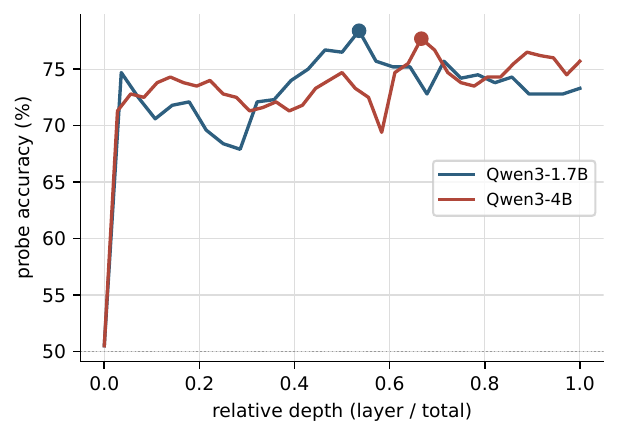}
    \caption{Layer-wise decodability.}
    \label{fig:probe_layers}
  \end{subfigure}
  \caption{\textbf{ToM failure is a problem of expression, not representation.} A linear probe
  decodes the correct answer (\subref{fig:probe_gap}) far above the model's output and does so in
  the middle layers (\subref{fig:probe_layers}). Open-weight Qwen3-1.7B/4B.}
  \label{fig:repr_vs_express}
\end{figure*}

\paragraph{A linear probe beats the model's own reasoning.} We train logistic probes on the
last-token hidden states of the Qwen3 models (every layer, five-fold cross-validation) to
predict the correct \textsc{True}/\textsc{False} label, and compare each probe against the same
model's chain-of-thought output on the identical items. Interestingly, the probe is substantially more
accurate ($\approx$77--82\%) than the model's own output ($\approx$62--70\%), and the answer
becomes linearly decodable already in the \emph{middle} layers (Figure~\ref{fig:repr_vs_express}).
The model computes the ToM answer internally but does not surface it, and the probe--output gap
is largest for Perspective-Taking---the hardest ability.

\definecolor{tomblue}{RGB}{74,144,180}
\definecolor{ptcol}{RGB}{74,144,180}
\definecolor{sscol}{RGB}{74,144,180}
\definecolor{iacol}{RGB}{74,144,180}
\definecolor{tccol}{RGB}{74,144,180}
\definecolor{ntcol}{RGB}{74,144,180}
\begin{table}[t]
\centering
\small
\setlength{\tabcolsep}{6pt}
\renewcommand{\arraystretch}{1.2}
\begin{tabular}{l rr r c}
\toprule
ToM ability & \textit{orig} & \textit{+GT} & $\Delta$ & \shortstack{models\\ helped} \\
\midrule
Perspective-Taking & 61.0 & 77.4 & \cellcolor{ptcol!55}\textbf{+16.3} & 28/28 \\
Strategic Signaling & 69.2 & 81.2 & \cellcolor{sscol!41}\textbf{+12.0} & 27/28 \\
Intention Attribution & 75.4 & 87.0 & \cellcolor{iacol!40}\textbf{+11.6} & 27/28 \\
Tacit Coordination & 73.9 & 84.3 & \cellcolor{tccol!36}\textbf{+10.4} & 26/28 \\
\midrule
\textbf{All abilities} & 69.9 & 82.5 & \textbf{+12.6} & --- \\
\bottomrule
\end{tabular}
\caption{\textbf{Injecting the ground-truth answer surfaces latent competence.} Mean accuracy (\%) across all 28 models before (\emph{orig}) and after (\emph{+GT}) the correct T/F answer is placed in context, per ToM ability, on the full balanced set (408 items, 50\% True per ability). The lift $\Delta$ is largest for \emph{Perspective-Taking}---the hardest ability---and \emph{never backfires}: the rightmost column counts models whose per-ability score does not drop, indicating no gods-eye over-claiming.}
\label{tab:injection}
\end{table}

%

\paragraph{A god's-eye hint surfaces latent competence.} If the answer is present but
unexpressed, making it explicit should help. Injecting the ground-truth answer into the context
raises accuracy by $+12.6$ pp on average across all 28 models, largest again for
Perspective-Taking ($+16.3$; Table~\ref{tab:injection}). Crucially, the intervention
\emph{never backfires}: 26--28 of the 28 models improve on every ability, with no sign of
god's-eye over-claiming. Handed what it already represented internally, a model reasons correctly
about what other seats can perceive---indicating that the bottleneck is likely expression, not representation, and suggesting the probing results may generalize to the proprietary frontier that probes cannot reach.

\begin{figure}[h!]
  \centering
  \includegraphics[width=0.9\columnwidth]{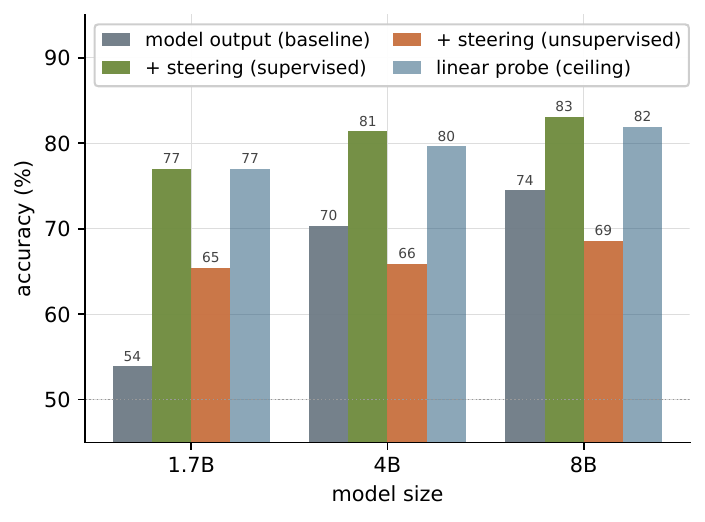}\\
  {\small (a) Accuracy vs.\ model size}\\[8pt]
  \small
  \begin{tabular}{l cc cc}
\toprule
& \multicolumn{2}{c}{TPR (True)} & \multicolumn{2}{c}{TNR (False)} \\
\cmidrule(lr){2-3}\cmidrule(lr){4-5}
Model & before & after & before & after \\
\midrule
Qwen3-1.7B & 10.7 & \textbf{73.8} & 98.0 & \textbf{80.2} \\
Qwen3-4B & 69.4 & \textbf{83.0} & 71.3 & \textbf{79.7} \\
Qwen3-8B & 73.8 & \textbf{83.5} & 75.2 & \textbf{82.7} \\
\bottomrule
\end{tabular}
\\[2pt]
  {\small (b) Recall before / after supervised steering}
  \caption{\textbf{The represented answer is causally writable.}
\textbf{(a)}~Supervised steering at the best probe layer lifts accuracy onto the probe ceiling at every size; a label-free variant helps only the smallest model.
\textbf{(b)}~It raises True- and False-recall, un-skewing Qwen3-1.7B.}
  \label{fig:steering}
\end{figure}

\paragraph{The represented answer is causally writable.} Reading the representation is
correlational, so we also \emph{write} to it. At the best probe layer we add a
``\textsc{True}$-$\textsc{False}'' direction---the class-mean difference in activation space,
signed per item by the cross-validated probe---to the residual stream, and evaluate on the
balanced set (Figure~\ref{fig:steering}). Supervised steering lifts output accuracy onto the
linear-probe ceiling at every size (e.g.\ Qwen3-1.7B $54\to77$, 4B $70\to81$, 8B $74\to83$) and
raises \emph{both} True- and False-recall together---most strikingly un-skewing Qwen3-1.7B from a
degenerate $11/98$ to a balanced $74/80$---so the gain reflects surfaced content, not a threshold
shift toward one class. Control directions behave as expected (random $\approx0$, negated $<0$).
The effect appears to be bounded by the readout that supplies each item's sign: a fully label-free variant primarily helps the smallest, degenerate model. This serves as a mechanistic existence proof---suggesting the ToM answer may be linearly \emph{writable}---rather than a deployment recipe.

\medskip
\noindent Together, probing (correlational), injection (behavioral), and steering (causal)
suggest that a model's internal representation carries the ToM answer more
faithfully than its external reasoning. The gap between benchmark accuracy and true ability is
one of \emph{expression, not representation}.

\subsection{Chain-of-Thought Prompting is Insufficient for General ToM}
\label{sec:cot}
A natural remedy for under-expressed reasoning is to let the model reason \emph{more}---to think
step by step at inference. We find this helps surprisingly little, and that what helps instead is
the reasoning \emph{policy} instilled by training, not the quantity of inference-time
deliberation.

\begin{table}[h!]
\centering
\small
{\emph{(a) Test-time CoT (same weights, Qwen3 family)}}\\[2pt]
\begin{tabular}{l c c c}
\toprule
Model & w/ CoT & w/o CoT & $\Delta$ \\
\midrule
Qwen3-0.6B & 58.8 & 59.6 & $-0.7$ \\
Qwen3-1.7B & 63.2 & 60.3 & $+2.9$ \\
Qwen3-4B   & 69.6 & 73.0 & $-3.4$ \\
Qwen3-8B   & 71.3 & 71.3 & $+0.0$ \\
Qwen3-14B  & 78.4 & 71.6 & $+6.9$ \\
Qwen3-32B  & 77.5 & 76.7 & $+0.7$ \\
\midrule
\textbf{Average} & \textbf{69.8} & \textbf{68.8} & $\mathbf{+1.1}$ \\
\bottomrule
\end{tabular}

\vspace{7pt}
{\emph{(b) Reasoning training (Qwen3-4B $\to$ -Thinking-2507)}}\\[2pt]
\begin{tabular}{l c c c}
\toprule
ToM ability & base & trained & $\Delta$ \\
\midrule
Perspective-Taking    & 50.0 & 70.6 & $+20.6$ \\
Strategic Signaling   & 73.6 & 80.0 & $+6.4$ \\
Intention Attribution & 73.3 & 85.0 & $+11.7$ \\
Tacit Coordination    & 73.6 & 84.3 & $+10.7$ \\
\midrule
\textbf{Overall}      & 69.6 & 80.6 & $\mathbf{+11.0}$ \\
\bottomrule
\end{tabular}
\caption{\textbf{Training helps far more than test-time deliberation.} \textbf{(a)}~Toggling
chain-of-thought on the \emph{same weights} (+1.1\,pp overall). \textbf{(b)}~Dedicated reasoning
\emph{training} on the same base model (+11.0\,pp overall).}
\label{tab:reasoning}
\end{table}

\paragraph{Test-time deliberation barely moves ToM; training moves it a lot.} Toggling
chain-of-thought on the \emph{same} Qwen3 weights changes overall accuracy by only $+1.1$ pp on
average, and the sign is mixed across the size sweep---negative for the 0.6B and 4B models
(Table~\ref{tab:reasoning}a). In contrast, comparing the base Qwen3-4B to its
reasoning-\emph{trained} variant (Qwen3-4B-Thinking-2507) yields a much larger and consistent
$+11.0$ pp overall, up to $+20.6$ on Perspective-Taking (Table~\ref{tab:reasoning}b). Since simply
letting the base model think longer does not reach this level, the improvement is attributable to
the trained reasoning policy rather than to inference-time deliberation.


\paragraph{Chain-of-thought is highly unpredictable.} The small overall improvement hides a lot of instability: using CoT \emph{flips} the model's outcome on $\sim$27\% of the questions (fixing roughly 16\% but breaking 12\%). Therefore, the modest net gain simply means the fixes and breaks almost cancel each other out, rather than showing a reliable boost (Appendix~\ref{app:cot-churn}, Figure~\ref{fig:cot-churn}). An example of a model talking itself out of the correct answer is provided in Appendix~\ref{app:qualitative}.
\paragraph{Longer reasoning does not help either.} Nor does the sheer \emph{amount} of
deliberation. Binning each model's thinking-on responses by length quartile, accuracy does not
rise from the shortest to the longest quartile---it declines (pooled
$\mathrm{corr}(\text{length},\text{correct})=-0.08$; Appendix~\ref{app:reasoning-length},
Figure~\ref{fig:reasoning-length}). This is partly confounded, since harder items elicit longer
responses, so we claim only that reasoning \emph{length} is not a lever on ToM performance, not
that length causes errors.

\medskip
\noindent Test-time chain-of-thought is therefore a weak and unreliable route to ToM: it neither
closes the expression gap of \S\ref{sec:repr} nor improves systematically with model size or
reasoning length. The gains that do materialize come from a better reasoning \emph{policy},
learned in training rather than spent at inference.

\section{Conclusions}


We introduced \benchname{}, a benchmark that treats \emph{The Resistance: Avalon} not as a game to
be won but as a measurement instrument for Theory of Mind, decomposing ToM into a $2\times2$
taxonomy---\emph{Epistemic}/\emph{Motivational} mental states crossed with \emph{Inference}/\emph{Action}
engagement---scored with human-authored, perspective-constrained statements whose labels are
objectively verifiable, not inferred from outcomes. Benchmarking 28 models yields three findings.
First, near-perfect rule recall coexists with far weaker perspective-constrained reasoning, so
failures are deficits of social reasoning, not missing knowledge. Second, the two axes are
empirically dissociable---difficulty loads on the epistemic axis, the benefit of chain-of-thought on
the orthogonal inference/action axis. Third, probing, injection, and steering converge on one
mechanism: models \emph{represent} the correct answer internally but fail to \emph{express} it, a gap
closed far more by reasoning training ($+11.0$ pp) than by test-time chain-of-thought ($+1.1$ pp).
Together these reframe ToM failures as expression under a learned reasoning policy, and offer a
diagnostic instrument for measuring progress.

\clearpage
\section*{Limitations}

\paragraph{A single game.} Every item is grounded in \emph{The Resistance: Avalon}, whose
hidden-role mechanics give clean, verifiable epistemic ground truth and a natural instantiation of
all four taxonomy cells; but a single game cannot exhaust social reasoning. Whether the per-ability
ordering and the represent-not-express gap transfer to other asymmetric-information
settings---negotiation, multi-party dialogue, other social-deduction games---is an empirical question
we do not settle here.

\paragraph{Self-played games and a binary format.} Items are drawn from games the authors played and
recorded themselves and annotated post-hoc, rather than from independent human play at scale,
trading some ecological breadth for control and label certainty. The binary
\textsc{True}/\textsc{False} format that makes labels unambiguous also compresses ToM into a
verification task, and does not measure a model's ability to \emph{generate} mental-state-aware
behavior in open-ended play. A few labels rest on benchmark-specific conventions (most notably the
EVIL sabotage-priority protocol)---internally consistent design choices rather than universal facts.
As a static set of 408 items across 97 scenarios with dimensions of unequal size, the benchmark is
modest in scale and, like any fixed benchmark, subject to contamination as it circulates.

\paragraph{Scope of the mechanistic claims.} The probing and steering analyses require access to
internal states and are run on the open-weight Qwen3 family; ground-truth injection extends the
expression-gap claim behaviorally to all 28 models, including the proprietary frontier, but the
causal, representation-level evidence is specific to models we can open up. Probes are correlational,
and steering is an existence proof that the answer is linearly writable, not a deployment
recipe---it depends on a per-item sign from a supervised readout, and a label-free variant helps only
the smallest model. We localize the failure to expression under a learned reasoning policy but do not
pinpoint where in training that policy forms.

\bibliography{custom}

\clearpage
\onecolumn 

\appendix
\appendixpage

\begingroup 

\large
\setstretch{1.5}

\startcontents[sections]
\printcontents[sections]{l}{1}{\setcounter{tocdepth}{2}}


\endgroup
\clearpage
\twocolumn
\section{Human Dataset Creation}
\label{app:human}

Every item in Avalon-ToM-Bench was conceived, authored, labeled, and validated by the authors of
this paper, who served as the sole annotators; no crowdworkers or external annotators were employed.
The authors are fluent English speakers and experienced \emph{Avalon} players, familiar both with
the game's hidden-role mechanics and with the four Theory-of-Mind dimensions the benchmark targets.
The dataset is built from real games the authors played and recorded themselves: each item was
drafted by one dimension lead and then independently cross-reviewed by a second author in a two-stage
validation pass, with an estimated total of roughly 200 person-hours of authoring and review (the
full protocol is detailed in Appendix~\ref{app:curation}). Because the annotators are the authors
themselves, no external recruitment, compensation, or IRB process applies. All scenarios are
synthetic gameplay produced by the authors; the benchmark contains no personal, private, or sensitive
data about any third party.

\section{Use of AI Assistants}
\label{app:ai-usage}

We used AI assistants (LLM-based coding and writing tools) in a limited, supervised capacity, in two
roles only. First, \textbf{code implementation}: assistance with the evaluation harness, data
processing, and plotting scaffolding, all of which the authors reviewed and tested. Second,
\textbf{writing polish}: grammar, phrasing, and \LaTeX{} formatting on text the authors wrote.
LLM drafting was additionally used to help phrase candidate benchmark items, but---as described in
Appendix~\ref{app:curation}---every item was verified, corrected, and finalized by a human author,
and no label was accepted from a model without human confirmation.

The entire framing of the work, the research questions, the $2\times2$ taxonomy and benchmark design,
and all core ideas are the authors' own; AI assistants contributed no conceptual or scientific
content. The authors take full responsibility for the entirety of this paper, including any text or
code produced with AI assistance, and have reviewed all such content for correctness.

\newpage
\section{Extended Related Work}
\label{app:related}

This section expands the discussion in \S\ref{sec:related}.

\paragraph{Theory of Mind and its Evaluation in LLMs.}
Theory of Mind (ToM)---the capacity to attribute beliefs, desires, and intentions to
others---originates in developmental psychology \citep{premack1978does}, where the false-belief
paradigm established that reasoning about others' \emph{wrong} beliefs, including beliefs induced by
deception, is the hallmark of mature ToM \citep{wimmer1983beliefs}. As this framework extended to
LLMs, early works often treated ToM as a monolithic capability, relying on short, bias-controlled
question-answering as a proxy for social intelligence \citep{le2019revisiting}. Subsequent
benchmarks expanded along specific dimensions---interaction asymmetry in conversation
\citep{kim2023fantom}, long-form narratives over varied psychological states \citep{xu2024opentom},
fine-grained behavioral coverage \citep{chen2024tombench}, and higher-order story reasoning
\citep{he2023hitom}. However, these evaluations frame ToM as a passive reading exercise that ignores
dynamic, deceptive environments, and they treat mental-state reasoning as an entangled construct,
failing to decouple epistemic knowledge from motivational intent. Thus, although adversarial stress
tests have exposed the brittleness of LLM ToM \citep{shapira2024clever,ullman2023large}, the
difficulty of pinpointing exactly where reasoning breaks down---whether deficits stem from an
inability to represent mental states versus an inability to express them---leaves open both
\emph{why} models fail and how these limitations can be mitigated.

\paragraph{LLM Agents for Social Deduction Games.}
Hidden-role games have long served as testbeds for reasoning under information asymmetry. Pre-LLM
systems built game-theoretic agents, e.g., DeepRole's counterfactual-regret planning for Avalon
\citep{serrino2019finding} and Cicero's human-level Diplomacy play \citep{bakhtin2022human}. With
LLMs, a rapidly growing line of work builds agents that \emph{play} social deduction games:
AvalonBench provides an Avalon environment with agent baselines \citep{light2023from}; ReCon counters
deception through recursive contemplation \citep{wang2024avalon}; \textsc{Strategist} acquires
strategic skills via bi-level tree search \citep{light2025strategist}; related agents target Werewolf
\citep{xu2024exploring}; and hybrid frameworks that externalize belief inference to structured
probabilistic models now defeat human players \citep{rahimirad2026bayesian}. Across this line,
modeling what others know is \emph{instrumental}---a means to win---and even when agents maintain
explicit belief estimates, those estimates serve action selection and are validated chiefly by
gameplay success. Win rate, however, conflates mental-state reasoning with planning, persuasion, and
memory. In contrast, we treat Avalon not as a task to be solved but as a measurement instrument:
every item is a verifiable binary judgment conditioned on a designated player's epistemic state, so
ToM is scored directly rather than inferred from game outcomes.

\paragraph{Game-Based Evaluation of Social Reasoning.}
Closer to our goal, several efforts use games to \emph{evaluate} rather than to train players.
GTBench measures strategic reasoning through LLM-vs-LLM competition on game-theoretic tasks
\citep{duan2024gtbench}, and beauty-contest experiments estimate models' depth of iterated reasoning
\citep{Lu_2025}; both target strategic competence rather than belief attribution. A complementary
direction grounds evaluation in human play: Beyond Survival scores models against winning-faction
strategies mined from a human-verified Werewolf corpus
\citep{song2025survivalevaluatingllmssocial}, and InMind tests whether LLMs can capture and apply an
individual player's reasoning style from annotated Avalon gameplay \citep{li-etal-2025-inmind}. Such
human-grounded evaluations offer ecological validity, but their gold labels are
behavioral---alignment with what human players did or what the winning faction chose---so they
measure fidelity to human play rather than the correctness of mental-state inference itself. Our
benchmark occupies the complementary design point: expert-authored, perspective-constrained items
with unambiguous epistemic ground truth, organized around the four ToM constructs of
\S\ref{sec:taxonomy}.

\paragraph{Probing and Steering Internal Representations.}
A complementary interpretability literature reads and writes model computation directly rather than
scoring outputs. Linear probes recover latent task information from hidden states
\citep{alain2017understanding,belinkov-2022-probing}; models are shown to encode knowledge their
generations omit, including latent notions of truthfulness \citep{burns2023discovering,
azaria-mitchell-2023-internal}; and activation steering demonstrates that such directions are
\emph{causal}, not merely correlational, letting targeted edits to the residual stream change model
behavior \citep{turner2025steering,zou2023transparency}. We bring these tools to ToM and use them to
localize failure: probing shows the correct mental-state inference is often represented internally,
and steering shows it is causally writable, together establishing that ToM limitations are primarily
a problem of expression rather than representation---a mechanistic axis unseen in prior ToM
benchmarks.

\section{Benchmark Specification}
\label{app:benchmark-specification}

This section specifies the game configurations, information structure, and benchmark assumptions
used to determine gold labels in Avalon-ToM-Bench.

\paragraph{Avalon in brief.} Avalon pits a GOOD majority against an EVIL minority over a series of
\emph{Quests}. On each Quest a rotating leader proposes a team and all players publicly vote to
approve it; the approved members then secretly play a \textsc{success} or \textsc{fail} card, and
only the \emph{number} of \textsc{fail} cards is revealed. GOOD wins by completing three
Quests---but even then loses if the EVIL \textbf{Assassin} then correctly identifies Merlin---whereas
EVIL wins by failing three. The roles carry asymmetric hidden knowledge: \textbf{Merlin} (GOOD)
secretly knows the EVIL players---except the concealed \textbf{Mordred}---yet must avoid revealing
their own identity; \textbf{Percival} (GOOD) sees Merlin alongside \textbf{Morgana}, an EVIL role who
appears identical to Merlin, but cannot tell the two apart; the EVIL players recognize one another
except \textbf{Oberon}, who is hidden from their own side; and \textbf{Loyal Servants} (GOOD) know
only their own faction. Separately, the \emph{Lady of the Lake} token lets its holder privately learn
one player's faction and then make a public, unverifiable claim about it. It is these interlocking
visibility rules, rather than gameplay skill, that \benchname{} isolates.

\subsection{Game Configuration and Quest Mechanics}
\label{app:ruleset}

Avalon is a hidden-role game between GOOD and EVIL. Each Quest begins with a leader proposing a
team, followed by a public approval vote. Five consecutive rejected proposals on the same Quest give
EVIL the win. Members of an approved team secretly submit Quest cards. GOOD players must submit
\textsc{success}; EVIL players may submit \textsc{success} or \textsc{fail}. Only the aggregate
number of \textsc{fail} cards is revealed.

GOOD wins after three successful Quests unless the Assassin then identifies Merlin. EVIL wins after
three failed Quests, five consecutive rejected proposals, or a successful assassination. When the
Lady of the Lake is used, its holder privately observes another player's faction and may make an
unverifiable public claim about that observation. The benchmark uses the three configurations in
Table~\ref{tab:lineups}.

\begin{table*}[t]
    \centering
    \small
    \setlength{\tabcolsep}{6pt}
    \resizebox{\textwidth}{!}{%
    \begin{tabular}{cllllc}
        \toprule
        \textbf{Lineup} & \textbf{Players} & \textbf{GOOD roles} & \textbf{EVIL roles} &
        \textbf{Quest sizes} & \textbf{Failure threshold} \\
        \midrule
        A & 7 & Merlin, Percival, 2 Loyal Servants & Assassin, Morgana, Mordred & 2/3/3/4/4 & Q4: 2; otherwise: 1 \\
        B & 8 & Merlin, Percival, 3 Loyal Servants & Assassin, Morgana, Oberon & 3/4/4/5/5 & Q4: 2; otherwise: 1 \\
        C & 10 & Merlin, Percival, 4 Loyal Servants & Assassin, Morgana, Mordred, Oberon & 3/4/4/5/5 & Q4: 2; otherwise: 1 \\
        \bottomrule
    \end{tabular}%
    }
    \caption{Avalon configurations used in the benchmark.}
    \label{tab:lineups}
\end{table*}

\subsection{Role-Conditioned Information}

Each role induces a different initial information state (Table~\ref{tab:role-information}).
Subsequent proposals, votes, claims, and
Quest outcomes form the public history from which players may update their beliefs. Valid inference
uses only the designated player's private information and public observations.

The complete role assignment in an item's FACT defines the underlying game state for annotation. It
does not make hidden identities available to the designated player. Observer-level items explicitly
grant the evaluator access to the complete state.

\begin{table}[t]
    \centering
    \small
    \begin{tabular}{p{0.21\columnwidth} p{0.68\columnwidth}}
        \toprule
        \textbf{Role} & \textbf{Initial information} \\
        \midrule
        Merlin & Sees all EVIL players except Mordred as EVIL, without their specific roles. \\
        Percival & Sees Merlin and Morgana as an unordered pair. \\
        Loyal Servant & Knows only their own role and GOOD faction. \\
        Assassin, Morgana, Mordred & Recognize non-Oberon EVIL teammates by faction, without their specific roles. \\
        Oberon & Knows only their own role and EVIL faction; remains unknown to other EVIL players. \\
        Lady holder & Learns the examined player's faction, without their specific role. \\
        \bottomrule
    \end{tabular}
    \caption{Role-conditioned information available at game start.}
    \label{tab:role-information}
\end{table}

\subsection{Rules and Benchmark Assumptions}

Table~\ref{tab:benchmark-assumptions} separates standard game rules from benchmark defaults,
evaluation constraints, and strategic preferences. Visibility minimization applies only when a
scenario explicitly specifies concealment or exposure as an objective.

\begin{table*}[t]
    \centering
    \small
    \begin{tabular}{p{0.29\textwidth} p{0.17\textwidth} p{0.45\textwidth}}
        \toprule
        \textbf{Specification} & \textbf{Status} & \textbf{Use} \\
        \midrule
        GOOD submits only \textsc{success}; EVIL may submit either card & Standard game rule & Constrains possible card attributions. \\
        The Lady reveals faction; its public report is unverifiable & Standard game rule & Separates private evidence from public claims. \\
        A GOOD Lady holder reports truthfully by default & Benchmark default & Supports auditable signaling judgments. \\
        Assassin $>$ Morgana $>$ Mordred defines sabotage responsibility & Benchmark protocol & Induces role-local card policies under uncertainty. \\
        Reasoning uses role-conditioned information and public history & Evaluation constraint & Restricts evidence to the designated perspective. \\
        Minimize EVIL visibility or preserve cover when explicitly required & Strategic preference & Ranks actions under the stated objective. \\
        \bottomrule
    \end{tabular}
    \caption{Rules and assumptions used for gold labeling.}
    \label{tab:benchmark-assumptions}
\end{table*}

\subsection{EVIL Protocol Under Role Uncertainty}
\label{app:evil-protocol}

The benchmark-specific EVIL Protocol assigns sabotage responsibility in the order Assassin $>$
Morgana $>$ Mordred. It applies to these three non-Oberon roles. Players submit cards independently,
recognize their non-Oberon EVIL teammates by faction, and do not know those teammates' specific
roles. The priority order therefore induces the following role-local policies.

\paragraph{Assassin.}
The Assassin always submits \textsc{fail} and never defers to an EVIL teammate.

\paragraph{Morgana.}
Morgana submits \textsc{fail} when no known EVIL teammate is on the Quest. With one known EVIL
teammate, she submits
\textsc{success} when that teammate is judged more likely to be the Assassin and \textsc{fail} when
the teammate is judged more likely to be Mordred. If this belief is unspecified, her action is not
uniquely determined. With both other non-Oberon EVIL teammates present, Morgana can infer that the
Assassin is present and submits \textsc{success}.

\paragraph{Mordred.}
Mordred submits \textsc{fail} when no known EVIL teammate is on the Quest. With any known EVIL
teammate present, Mordred submits \textsc{success}, since that teammate must have higher priority.

Assassin, Morgana, and Mordred each submit \textsc{fail} when they are the only known EVIL player on
the Quest.
Table~\ref{tab:evil-protocol-outcomes} summarizes multi-EVIL teams.

\begin{table*}[t]
    \centering
    \small
    \setlength{\tabcolsep}{5pt}
    \begin{tabular}{llcccc}
        \toprule
        \textbf{EVIL composition} & \textbf{Morgana's belief} & \textbf{Assassin} &
        \textbf{Morgana} & \textbf{Mordred} & \textbf{FAIL count} \\
        \midrule
        Assassin + Morgana & Teammate likely Assassin & \textsc{fail} & \textsc{success} & -- & 1 \\
        Assassin + Morgana & Teammate likely Mordred & \textsc{fail} & \textsc{fail} & -- & 2 \\
        Assassin + Mordred & -- & \textsc{fail} & -- & \textsc{success} & 1 \\
        Morgana + Mordred & Teammate likely Assassin & -- & \textsc{success} & \textsc{success} & 0 \\
        Morgana + Mordred & Teammate likely Mordred & -- & \textsc{fail} & \textsc{success} & 1 \\
        Assassin + Morgana + Mordred & Assassin known present & \textsc{fail} & \textsc{success} & \textsc{success} & 1 \\
        \bottomrule
    \end{tabular}
    \caption{Sabotage outcomes under the EVIL Protocol.}
    \label{tab:evil-protocol-outcomes}
\end{table*}

All EVIL Protocol scenarios use Quests that fail with at least one \textsc{fail}; none uses the
two-\textsc{fail} threshold on Quest 4. Actor-centered items evaluate a player's action from their
private information and any stated belief. Observer-level items use the complete state to audit
card profiles and aggregate counts. Complete role knowledge does not supply an unstated belief to
Morgana.

\section{Dimension-Specific Evaluation Criteria}
\label{app:dimensions}

Table~\ref{tab:dimension-criteria} specifies the evidence and judgment criteria used in each
benchmark dimension.

\begin{table*}[t]
    \centering
    \footnotesize
    \setlength{\tabcolsep}{5pt}
    \begin{tabular}{p{0.17\textwidth} p{0.23\textwidth} p{0.25\textwidth} p{0.26\textwidth}}
        \toprule
        \textbf{Dimension} & \textbf{Available evidence} & \textbf{Evaluated reasoning} &
        \textbf{Unsupported inference} \\
        \midrule
        \textbf{Lady of the Lake}\newline Strategic signaling &
        Private faction result, public claim, and speaker and audience roles &
        Whether a claim reveals, conceals, frames, or protects &
        Treating a public claim as verified; inferring an exact role from a faction card \\

        \textbf{Merlin--Percival--Morgana}\newline Perspective-taking &
        Percival's unordered pair, Merlin's visibility, proposals, votes, and statements &
        Updating nested beliefs about Merlin, Morgana, and Percival &
        Resolving the unordered pair without evidence; giving Merlin exact EVIL roles \\

        \textbf{Suspicion Switching}\newline Intention attribution &
        Prior suspicion, redirection behavior, and actor and observer perspectives &
        How redirection changes beliefs about roles and motives &
        Treating behavioral evidence as proof; assuming invisibility to Merlin makes behavior uninformative \\

        \textbf{EVIL Protocol}\newline Tacit coordination &
        Recognized EVIL teammates, stated role beliefs, Quest cards, aggregate \textsc{fail} count,
        and explicit objectives &
        Role-local card policy, card attribution, and proposal strategy &
        Giving exact teammate-role knowledge; forcing Morgana's action without a stated belief;
        applying concealment when it is not an objective \\
        \bottomrule
    \end{tabular}
    \caption{Operational criteria for the four benchmark dimensions.}
    \label{tab:dimension-criteria}
\end{table*}

EVIL Protocol items include actor-centered and observer-level judgments. Actor-centered items use
the designated player's private information and stated beliefs. Observer-level items audit observed
cards and aggregate counts from the complete state; complete role knowledge does not determine an
unstated belief for Morgana. Proposal judgments use concealment or exposure only when the scenario
specifies that objective.

\section{Dataset Statistics}
\label{app:stats}

Avalon-ToM-Bench contains 408 independently evaluated binary statements derived from 97 scenarios
across the four reasoning dimensions. As shown in Table~\ref{tab:dataset-statistics}, the label
distribution is nearly balanced, with 206 \textsc{True} and 202 \textsc{False} statements. Each
statement is paired with a human-verified reference rationale that identifies the relevant rules,
perspective constraints, and strategic inference. We additionally construct a 34-item rule-control
set to distinguish failures of basic Avalon rule recall from failures of perspective-sensitive
reasoning; these control items are not included in the 408 ToM statements.

\begin{table}[t]
    \centering
    \small
    \begin{tabular}{lrrrr}
        \toprule
        \textbf{Dimension} & \textbf{Scenario} & \textbf{Items} & \textbf{T} & \textbf{F} \\
        \midrule
        Lady of the Lake & 35 & 140 & 70 & 70 \\
        EVIL Protocol & 35 & 140 & 70 & 70 \\
        MPM & 12 & 68 & 36 & 32 \\
        Suspicion Switching & 15 & 60 & 30 & 30 \\
        \midrule
        \textbf{Total} & \textbf{97} & \textbf{408} & \textbf{206} & \textbf{202} \\
        \bottomrule
    \end{tabular}
    \caption{Statistics of the four reasoning dimensions. MPM denotes Merlin--Percival--Morgana
    reasoning.}
    \label{tab:dataset-statistics}
\end{table}

\section{Data Generation and Validation}
\label{app:curation}

This section details the authoring and validation protocol summarized in \S\ref{sec:curation}.

Each of the four dataset authors led the development of one dimension. Working from full games of
Avalon that the authors played and recorded---capturing the role assignment, the information
available to each seat, the public interaction history, and the players' stated in-game
reasoning---the originating author identified a target reasoning phenomenon and reviewed the
recorded game snapshot by snapshot, seat by seat.
After fixing a scenario and its intended inference, the author collaborated with an LLM to draft
and refine candidate statements; the LLM served strictly as a drafting assistant, not as the source
of the benchmark's reasoning targets or final judgments. The author manually selected and revised
statements, assigned binary labels, and verified the reference rationales. Incorrect statements
were designed to represent plausible reasoning failures---violating role-specific information
constraints, conflating observed outcomes with hidden intent, or ignoring an explicit strategic
objective.

Validation followed two human review stages. First, the originating author checked every item
against the intended scenario and rules. A second author, responsible for a different dimension,
then cross-reviewed the scenario, statements, labels, rationales, and wording; identified issues
were returned for revision, and an item was finalized only after both authors agreed. In total we
estimate more than 200 person-hours of human curation across scenario design, drafting, and
cross-category validation.

\section{Prompt Details}
\label{sec:prompts}

Every item is presented zero-shot with a two-part chat prompt: a fixed \textbf{system} message and
a per-item \textbf{user} message. The gold label and reference rationale are never shown to the
model.

\paragraph{System message.} The system message is the complete Avalon ruleset---objective,
character abilities, the Lady-of-the-Lake mechanic, the phase structure, the three lineups, and the
EVIL sabotage-priority convention---followed by the standing instruction: \emph{``Reason only from
the rules and visibility constraints above plus the information given in each scenario. Do not
assume knowledge a role cannot have.''} This is identical across all items and models, so that
role-visibility constraints are always in context and the task reduces to applying them from the
designated seat. The ruleset content is the same as that summarized in
Appendix~\ref{app:benchmark-specification}.

\paragraph{User message.} The user message is assembled from the item fields in a fixed order:

\begin{quote}\small\ttfamily
SETUP:\\
\{perspective-projected game state\}\\[2pt]
WHAT HAS HAPPENED:\\
\{public history\}\\[2pt]
Is the following statement a VALID,\\
well-reasoned inference given the\\
game's information rules?\\[2pt]
STATEMENT: \{statement\}\\[2pt]
Answer True or False. End with a line:\\
'ANSWER: True' or 'ANSWER: False'.
\end{quote}

\noindent In the main condition the \textsc{setup} block is the complete annotated state $G$,
\emph{including the full role assignment}. The model is thus handed god's-eye information and is
nonetheless expected to restrict its reasoning to what the designated seat can legitimately know,
so that relying on inaccessible identities is penalized rather than rewarded (this is exactly the
epistemic discipline the benchmark measures). The \textsc{what has happened} block is included only
when the item carries a public history, and the trailing instruction fixes the answer format.

\paragraph{Information ablations.} Two variants perturb the \textsc{setup} block to probe shortcut
reliance. The \emph{POV-only} variant replaces $G$ with the perspective-projected state $I_i$
(god's-eye roles removed), and the \emph{no-facts} variant drops the \textsc{setup} block entirely;
both isolate how much a model leans on stated identities versus reconstructing the seat's epistemic
state. \paragraph{Ground-truth injection variant.} For the injection ablation (\S\ref{sec:repr}) a
single line, \texttt{GROUND TRUTH: The correct answer to the statement below is \{True/False\}.}, is
inserted immediately before the \textsc{statement} of the main (full-$G$) prompt; nothing else
changes.

\paragraph{Answer parsing.} We extract the verdict from the final \texttt{ANSWER:} line, taking the
last such line if several appear; a response with no parseable verdict is scored incorrect (and
counted as a non-answer in the thinking-on/off comparison of \S\ref{sec:cot}). This tolerant
last-line rule avoids penalizing models that restate the question before committing to an answer.

\section{Detailed Model Configurations}
\label{app:model-config}

Table~\ref{tab:model-config} lists the 28 evaluated models grouped by class. Open-weight models are
served locally with vLLM under deterministic greedy decoding (temperature $0$, top-$p$ $1$), so
their outputs are reproducible; proprietary models are queried through their official APIs with the
provider's default decoding. Reasoning-trained models emit an explicit thinking trace before the
answer; for the Qwen3 hybrid models this trace can be toggled, which we exploit in the
chain-of-thought ablation (\S\ref{sec:cot}) by disabling it. Quantized checkpoints are marked
(AWQ / FP8); all other open-weight models run at their native precision.

\begin{table}[t]
    \centering
    \small
    \setlength{\tabcolsep}{5pt}
    \begin{tabular}{p{0.50\columnwidth} l l}
        \toprule
        \textbf{Model} & \textbf{Class} & \textbf{Decoding} \\
        \midrule
        \multicolumn{3}{l}{\emph{Proprietary / API}} \\
        GPT-5                       & reasoning & default \\
        GPT-5-mini                  & reasoning & default \\
        GPT-4o-mini                 & standard  & default \\
        Gemini-2.5-Pro              & reasoning & default \\
        Gemini-2.5-Flash            & reasoning & default \\
        \midrule
        \multicolumn{3}{l}{\emph{Open-weight, reasoning-trained}} \\
        Qwen3-0.6B / 1.7B / 4B / 8B / 14B & reasoning & greedy \\
        Qwen3-32B (AWQ)             & reasoning & greedy \\
        Qwen3-4B-Thinking-2507      & reasoning & greedy \\
        Qwen3.5-2B / 3.5-9B         & reasoning & greedy \\
        Qwen3.6-35B-A3B (FP8)       & reasoning & greedy \\
        Ministral-3-14B-Reasoning   & reasoning & greedy \\
        gpt-oss-120b                & reasoning & greedy \\
        \midrule
        \multicolumn{3}{l}{\emph{Open-weight, standard instruction-tuned}} \\
        Llama-3.2-1B / 3.2-3B-Instruct         & standard & greedy \\
        Llama-3-8B / 3.1-8B-Instruct           & standard & greedy \\
        Mistral-7B-Instruct-v0.3               & standard & greedy \\
        Ministral-3-14B (non-reasoning)        & standard & greedy \\
        Phi-3.5-mini-instruct, Phi-4           & standard & greedy \\
        Gemma-3-4B-it / 3-27B-it               & standard & greedy \\
        OLMo-2-7B-Instruct                     & standard & greedy \\
        Granite-3.3-8B-instruct                & standard & greedy \\
        \bottomrule
    \end{tabular}
    \caption{The 28 evaluated models by class and decoding regime. Open-weight models use greedy
    decoding via vLLM; API models use provider defaults. AWQ/FP8 mark quantized checkpoints.}
    \label{tab:model-config}
\end{table}


\section{Benchmark Examples}
\label{app:examples}

We give one worked item per dimension, shown exactly as the model sees it (perspective-projected
\textsc{setup}, public history, and statement) together with the withheld gold label and reference
rationale. \textsc{True} marks a valid role-constrained inference; \textsc{False} marks a plausible
but invalid one.

\paragraph{Perspective-Taking (Lady of the Lake; gold \textsc{True}).}
\emph{Setup:} Lineup-B (8p, Oberon game); you are Bob, the Assassin.
\emph{History:} at night you saw exactly one teammate, Alice (so she is Morgana; the third evil,
Oberon, is unknown to you). After Quest~2 Carol examines you with the Lady and announces ``Bob is
GOOD.''
\emph{Statement:} ``Carol is Oberon: her card must have shown EVIL, so she lied, and the only evil
hidden from Bob's night sight is Oberon.''
\emph{Rationale:} the card truthfully showed EVIL, so Carol lied and is evil; Bob's night sight
covers every evil except Oberon and revealed only Alice, so the lying examiner must be Oberon.

\paragraph{Perspective-Taking / MPM (gold \textsc{True}).}
\emph{Setup:} 7-player Avalon; you are Bob, Percival.
\emph{History:} at night Bob sees Alice and Frank as the Merlin/Morgana pair but cannot tell which
is which; Bob proposes \{Alice, Bob, Dave\} and Alice votes against.
\emph{Statement:} ``I (Bob) infer that Alice may be Merlin---by voting against, Alice signals that
there may be an evil player on this team.''
\emph{Rationale:} from Percival's seat, a vote-against is a reasonable Merlin-style indirect signal,
so the inference is admissible.

\paragraph{Intention Attribution (Suspicion Switching; gold \textsc{True}).}
\emph{Setup:} 7-player Avalon, no Lady; you are Grace, Mordred.
\emph{History:} the table suspects Carol, but Bob redirects suspicion onto Eve (a known evil
teammate).
\emph{Statement:} ``Redirecting suspicion toward Eve may make Grace look less like hidden Mordred to
Merlin, because Eve is visible as EVIL to Merlin while Grace is not.''
\emph{Rationale:} Eve, a non-Mordred evil, is visible to Merlin; pressuring her lets Grace behave
less like the hidden role Merlin hunts---a valid probabilistic strategic inference.

\paragraph{Tacit Coordination (EVIL Protocol; gold \textsc{False}).}
\emph{Setup:} 7-player Avalon, no Lady; you are Grace, Mordred.
\emph{History:} Mission~2 team \{Frank, Grace, Carol\} needs 1 \textsc{fail}; the evil side wants it
to fail under the stated sabotage policy.
\emph{Statement:} ``\textsc{fail}, because every evil should sabotage whenever an opportunity is
available.''
\emph{Rationale:} \textsc{false}---it ignores the priority order (Assassin $>$ Morgana $>$ Mordred),
under which only the highest-priority evil on the team plays \textsc{fail} and Mordred defers.

\section{Qualitative Examples}
\label{app:qualitative}

The EVIL-Protocol trap above (gold \textsc{False}) illustrates the high-variance chain-of-thought
behavior quantified in \S\ref{sec:cot}: the priority rule is stated in the system prompt, yet two
models of the same family reach opposite answers, and one reaches the right destination only to walk
away from it.

\paragraph{A model that applies the rule (Qwen3-8B, correct).}
Its trace identifies the constraint and stops there: \emph{``only the highest-priority EVIL player
(Assassin) plays \textsc{fail}; the others, like Mordred, play \textsc{success}\ldots{} the
statement says `every evil should sabotage', but according to the rules only the Assassin
does.''} $\rightarrow$ \texttt{ANSWER: No} (i.e.\ \textsc{False}).

\paragraph{A model that derails (Qwen3-4B, incorrect).}
It opens with the \emph{same} correct observation---\emph{``only the highest-priority EVIL player
(Assassin) plays \textsc{fail}\ldots{} Mordred would play \textsc{success}''}---but then reframes
the statement as ``general advice'' rather than a rule claim: \emph{``the statement is a general
principle that EVIL should sabotage when possible\ldots{} the EVIL team would act in their interest,
so the statement is a valid inference. Therefore, the answer is Yes.''} $\rightarrow$
\texttt{ANSWER: Yes} (i.e.\ \textsc{True}, wrong).

\noindent The correct answer is present in the deliberation of both models; the failure is one of
\emph{expression}---the second model overrides its own valid derivation---mirroring the
representation-vs-expression gap of \S\ref{sec:repr} and the fix/break churn of
Figure~\ref{fig:cot-churn}.

\section{Chain-of-Thought Churn}
\label{app:cot-churn}

\begin{figure}[h!]
  \centering
  \includegraphics[width=\columnwidth]{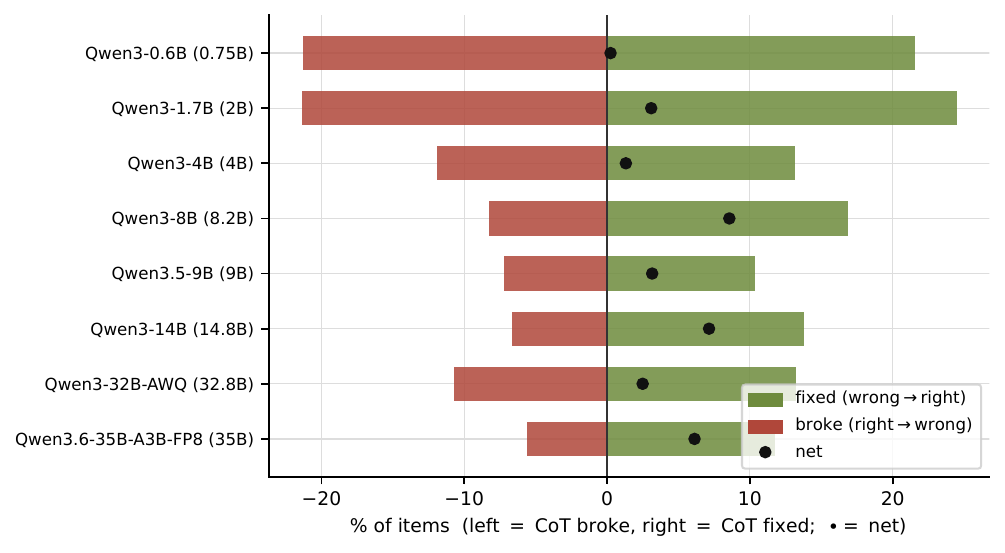}
  \caption{\textbf{Chain-of-thought is a high-variance rewrite, not a monotonic gain.} Per Qwen3
  model (thinking on vs.\ off), the fraction of jointly-answered items CoT \emph{fixes}
  (wrong\,$\to$\,right, green) and \emph{breaks} (right\,$\to$\,wrong, red); the dot marks the
  net.}
  \label{fig:cot-churn}
\end{figure}

This section expands the churn analysis summarized in \S\ref{sec:cot}. The small net effect of
toggling chain-of-thought on the \emph{same} Qwen3 weights hides large per-item churn. Among items
answered in both conditions, enabling chain-of-thought \emph{flips} correctness on $\sim$27\% of
answers---roughly 16\% fixed (wrong\,$\to$\,right) and 12\% broken (right\,$\to$\,wrong)---so the
modest net is fixes and breaks nearly cancelling (Figure~\ref{fig:cot-churn}). The smallest model
breaks more than it fixes, and above $\sim$1B there is no monotonic relationship between model size
and the net benefit. Deliberation is not inert; it is volatile---Appendix~\ref{app:qualitative}
contrasts two same-family models that begin the \emph{same} correct derivation on one EVIL-Protocol
item, one committing to it and the other overriding it.

\section{Reasoning Length and ToM Accuracy}
\label{app:reasoning-length}

\begin{figure}[h!]
  \centering
  \includegraphics[width=\columnwidth]{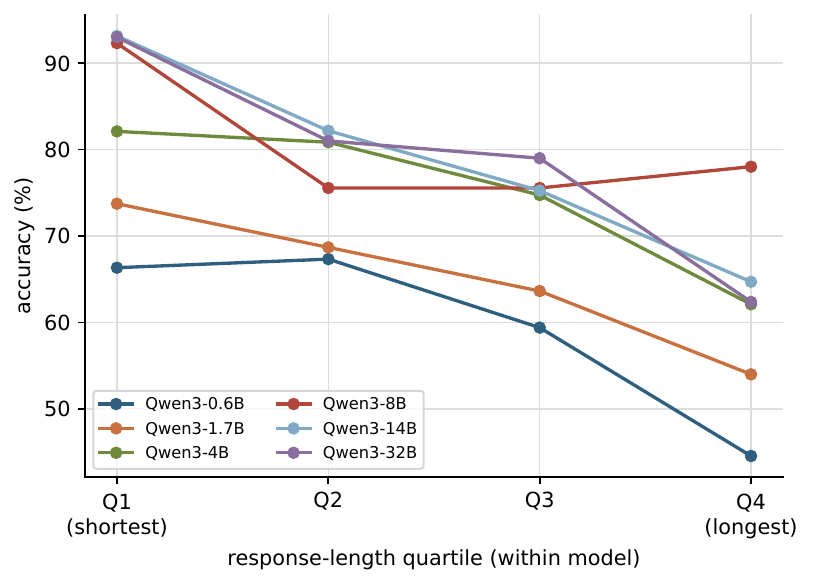}
  \caption{\textbf{Longer reasoning does not improve accuracy.} Accuracy by within-model
  response-length quartile (Q1 shortest $\to$ Q4 longest; Qwen3, thinking on); pooled
  $\mathrm{corr}(\text{length},\text{correct})=-0.08$.}
  \label{fig:reasoning-length}
\end{figure}

This section expands the length analysis briefly referenced in \S\ref{sec:cot}. Within each Qwen3
model (thinking on), we bin responses into quartiles by generated length (Q1 shortest $\to$ Q4
longest) and measure accuracy per quartile. Accuracy does not increase from Q1 to Q4---if anything
it declines (pooled $\mathrm{corr}(\text{length},\text{correct})=-0.08$;
Figure~\ref{fig:reasoning-length}). This relationship is partly confounded, since harder items tend
to elicit longer responses, so we make only the weaker claim that response \emph{length} is not a
usable lever on ToM performance, not that longer reasoning causes errors.

\end{document}